\documentclass[letterpaper, 10 pt, conference]{ieeeconf}  

\usepackage{booktabs}

\usepackage{fancyhdr}

\usepackage{fancyhdr}
\usepackage{lastpage}

\fancypagestyle{firstpage}{
  \fancyhf{}
  \fancyhead[C]{\footnotesize Accepted for publication in the proceedings of the \textit{2026 IEEE 9th International Conference on Soft Robotics (RoboSoft)}}
  \fancyfoot[C]{\scriptsize \copyright~2026 IEEE. Personal use of this material is permitted. Permission from IEEE must be obtained for all other uses, in any current or future media, including reprinting/republishing this material for advertising or promotional purposes, creating new collective works, for resale or redistribution to servers or lists, or reuse of any copyrighted component of this work in other works.}

}

\IEEEoverridecommandlockouts                              

\usepackage{cite}
\usepackage{amsmath,amssymb,amsfonts}
\usepackage{hyperref}
\usepackage{cleveref}
\usepackage{graphicx}
\usepackage{algorithmic}
\usepackage{graphicx}
\usepackage{textcomp}
\usepackage{colortbl}
\usepackage[table,xcdraw,dvipsnames]{xcolor}

\newcommand{\remove}[1]{}

\begin{document}

\title{\LARGE \bf
ZipFold: Modular Actuators for Scaleable Adaptive Robots
}

\author{
Niklas Hagemann and
Daniela Rus
\thanks{N. Hagemann and D. Rus are with the Computer Science and Artificial Intelligence
Laboratory, MIT, Cambridge,
MA 02139 USA, 
    {\tt\small \{hagemann, rus\}@csail.mit.edu}}%
}

\maketitle

\thispagestyle{firstpage}  
\pagestyle{empty}

\begin{abstract} 

There is a growing need for robots that can change their shape, size and mechanical properties to adapt to evolving tasks and environments. However, current shape-changing systems generally utilize bespoke, system-specific mechanisms that can be difficult to scale, reconfigure or translate from one application to another. This paper introduces a compact, easy-to-fabricate deployable actuator that achieves reversible scale and stiffness transformations through compound folding and zipping of flexible 3D-printed plastic strips into square-section deployable beams. The simple actuation method allows for smooth, continuous transitions between compact (flexible) and expanded (quasi-rigid) states, facilitating diverse shape and stiffness transformations when modules are combined into larger assemblies. The actuator’s mechanical performance is characterized and an integrated system involving a four-module adaptive walking robot is demonstrated.
\end{abstract}
\section{Introduction}
\label{sec:introduction}

Robots that can adapt their shape or mechanical properties in response to changing tasks and environments are critical for applications ranging from space exploration, autonomous inspection, search-and-rescue, to modular or morphing robotic systems \cite{polzin_robotic_2025, yim_modular_2007, liu_reconfiguration_2019, zhao_modular_2025}. In particular, compact robots that can be deployed into confined spaces and later expand, or reconfigure in order to perform a range of mechanical tasks (e.g. inspection inside of an aircraft interior, industrial plant, or collapsed environments), remain a key unmet challenge.

\begin{figure}[t]
\centering
\includegraphics[width=1\linewidth]{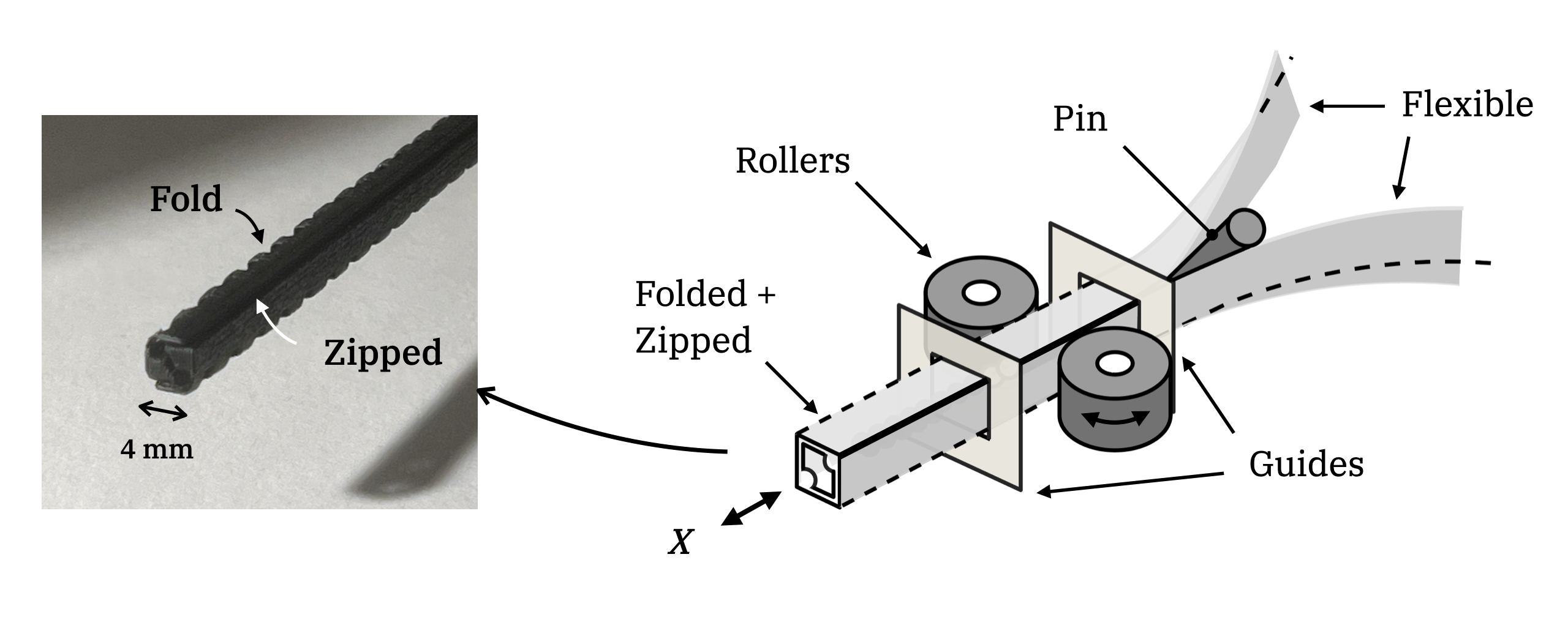}
\caption{\textbf{The square Fold-and-Zip principle.} Coupled counter-rotating rollers pull the flat (flexible) zipper strips through two guides, folding and zipping the strips together in a single continuous step and deploying a square-section beam. A simple passive pin ensures reversibility by unzipping the strips when the rollers are driven in reverse.}
\label{fig:actuation_principle}
\end{figure}

A class of systems that fundamentally addresses scale-change and transformation is the field of deployable structures, which has advanced significantly under the constraints of space applications \cite{crawford_strength_1971}. Space missions impose stringent requirements: large systems like solar arrays or telescopes must be lightweight and compactly packaged during launch, but functionally rigid once deployed \cite{miura_forms_2020}. These challenges have inspired a range of transformation strategies, from coil-able tape-measure–like systems \cite{thomson_deployable_1994, fernandez_advanced_2017} to origami-inspired designs that unfold \cite{miura_forms_2020}. Beyond aerospace, origami folding principles have been applied in domains like minimally invasive medical devices and compact robotic systems \remove{, where reduced weight, reversibility, and easily deployable mechanisms are equally valuable}. These examples demonstrate the utility of geometric- and material-based strategies for shape transformation, yet many solutions remain complex to fabricate, rely on exotic engineering materials and tuned towards highly domain-specific performance requirements (e.g. deployment of solar arrays in zero-gravity), resulting in highly directional mechanical characteristics, prone to failure under less controlled loading conditions \cite{fernandez_advanced_2017}. There is a need for large aspect-ratio deployable actuators that are simply designed and easy to combine and integrate into a range of deployable and shape- or stiffness-adaptive robotic systems. To address this gap, we here propose a new class of compact deployable actuator that can enable reversible shape- and stiffness-change through compound folding and zipping of flexible planar strips (Figure \ref{fig:actuation_principle}), offering a pathway toward highly scalable adaptive soft-robotic systems with distributed shape and stiffness adaptability.

\begin{figure}[!t]
\centering
\includegraphics[width=0.95\linewidth]{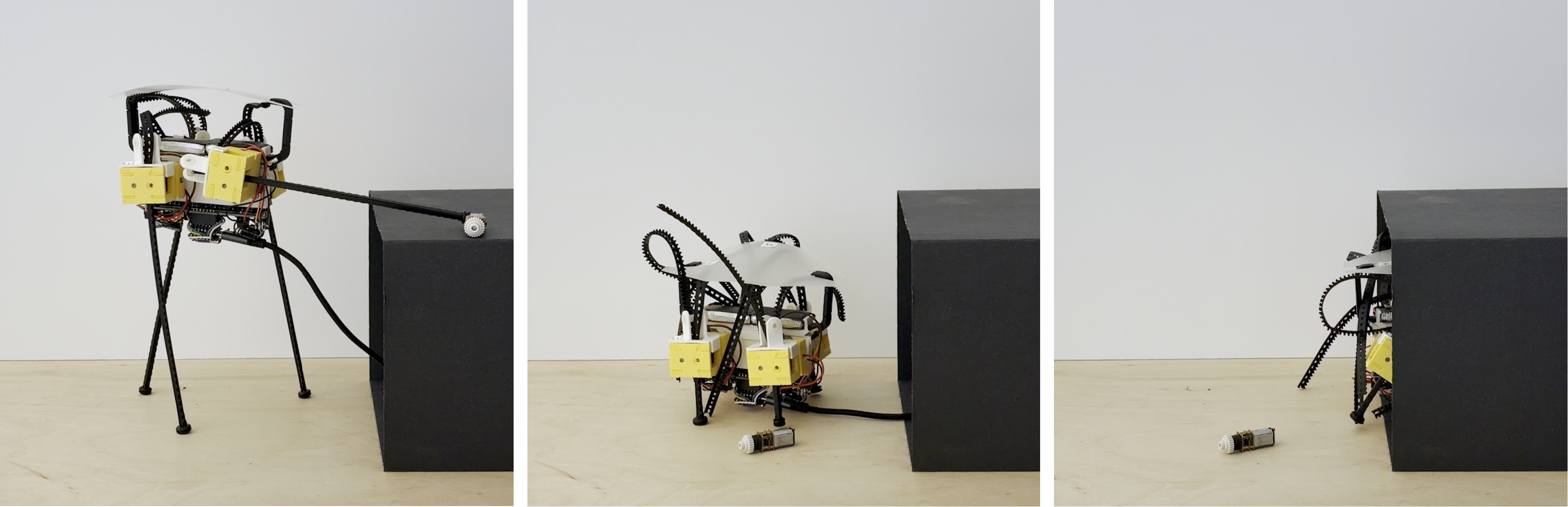}
\caption{\textbf{Walking robot composed of four ZipFold modules}. The robot can expand and contract to adapt to changes in its environment (pictured: expanded, crouching and walking into a confined space).}
\label{fig:four_module_robot}
\end{figure}

\begin{figure}[t]
\centering
\includegraphics[width=0.95\linewidth]{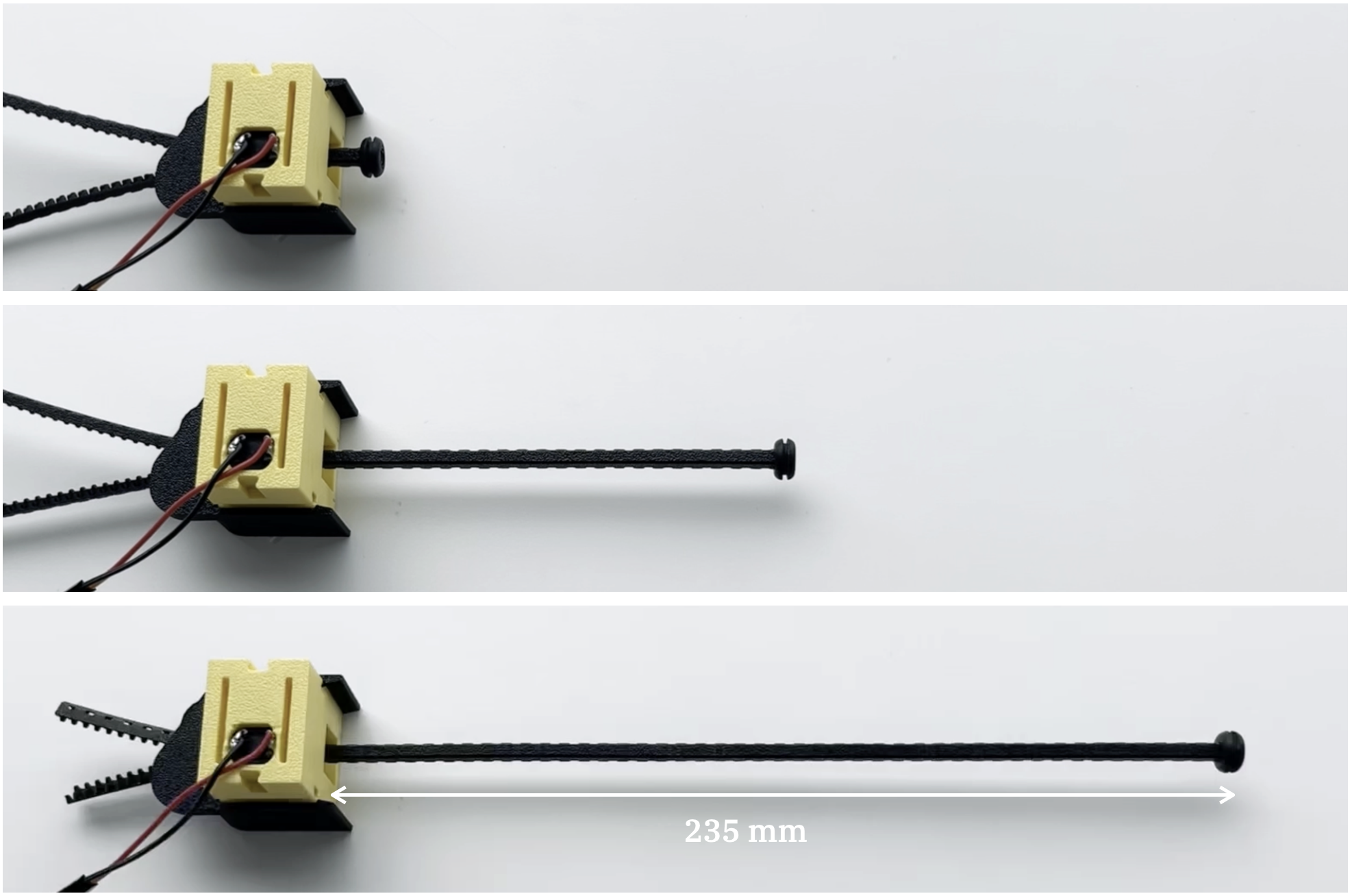}
\caption{\textbf{Extension of a single ZipFold actuator.} Two flexible strips are folded along their longitudinal axis and zipped at their edges to form a square-section deployable beam.}
\label{fig:deployment}
\end{figure}

\begin{figure}[t]
\centering
\includegraphics[width=1\linewidth]{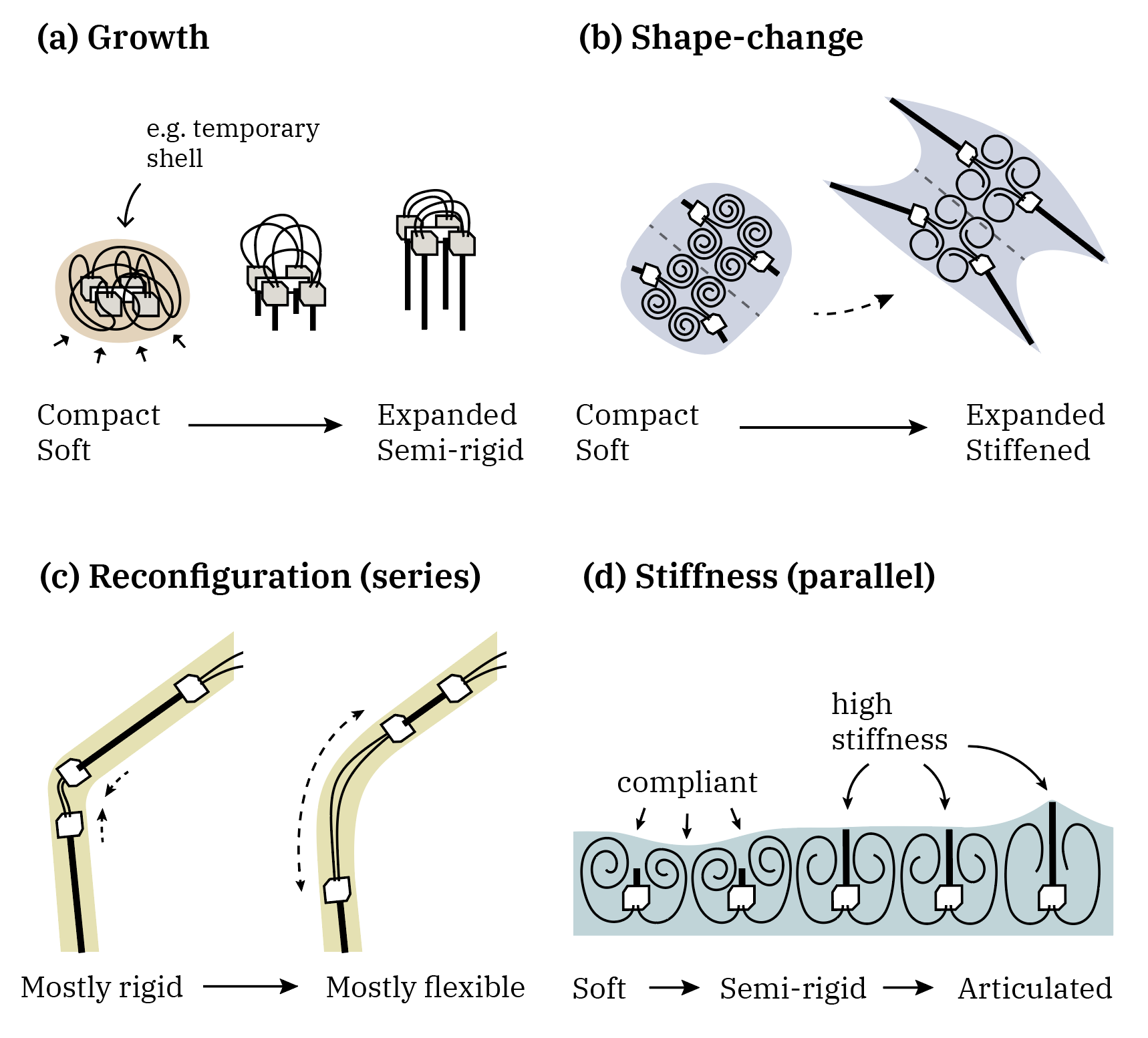}
\caption{\textbf{Application areas for ZipFold modules.} Modules could be combined to form a variety of adaptive soft-rigid systems: (a) a deployable, impact-resistant robot that starts off soft, (b) a soft adaptive (e.g. aquatic) robot, (c) adjustable joints within a hybrid soft/rigid robot or soft exoskeleton, (d) scaled-down modules as actuators within a smart articulated surface.}
\label{fig:applications}
\end{figure}



 Large transformations in scale have generally been achieved via two strategies: (1) folding (e.g. chains of origami- or scissor-based linkages)  \cite{melancon_multistable_2021, ou_kinetix_2018, filipov_origami_2015, bertoldi_flexible_2017, fu_droppop_2025} and (2) coiling  \cite{miura_forms_2020}\cite{coad_vine_2020}. Space-based applications regularly utilize both, with tape-measure like deployable booms acting as actuators and supports for unfolding solar arrays, which need to be packed tightly during launch and then unfolded in orbit \cite{miura_forms_2020}. In a robotics context, deployable boom-based systems have been proposed for compact exploration robots that can be lowered into caves and maneuver themselves in a tensile manner by expanding and pulling themselves through an internal space \cite{morton_task-driven_2024}. Coiling has been widely used as a strategy for compact storage, but helically interlocking (zipped) coils have also been used in a deployed manner to address the compressive load limitations of regular boom-based systems: with helical steel zipper systems finding their way into theaters as platforms that can emerge out of a stage to lift an actor or prop \cite{noauthor_spiralift_nodate}. Roboticists have applied similar concepts to develop modular, reconfigurable truss systems composed of helical zipped plastic coils \cite{collins_design_2016} \cite{kim_highly_2023}.

Large deployable and shape-changing robotic systems have faced several key limitations in the past. Generally originating from very particular sets of engineering constraints, e.g. such as the packing requirements of space missions, they are often highly bespoke, consisting of complex mechanical assemblies, and utilizing high-performance materials like carbon-fiber composites or steel coils \cite{fernandez_advanced_2017}. Many systems, particularly those reliant on folding mechanisms, are designed with at most two states in mind (stowed and deployed), lacking the capacity for continuous control in-between these configurations and in many cases conceived only as one-way (non-reversible) deployments \cite{hedayati_designing_2022}. Furthermore, many designs are highly directional in their performance characteristics (e.g. by being designed for the gravity-less environment of space) or depend on tensioning mechanisms to achieve functional stiffness for manipulation tasks that involve bending moments or larger compressive loads \cite{collins_design_2016}.

To address these shortcomings and bring insights from the field of deployable systems into the realm of soft and adaptive robotics, we present ZipFold: a compact, easy-to-fabricate deployable actuator that allows for smooth, reversible transitions between stowed (flexible) and expanded (semi-rigid) states. The actuators are inherently modular and can utilize arbitrary lengths of flexible 3D-printed strips that can be reversibly folded and interlocked (zipped) together in a single continuous actuation step (see Figure \ref{fig:actuation_principle}), to form semi-rigid deployable beams (see Figure \ref{fig:deployment}). The module design is simple to fabricate, compact and scalable: multiple actuators can be combined in parallel or series to construct shape-, scale-, and stiffness-changing robotic systems with localized control over shape and stiffness, enabling a range of possible applications (Figure \ref{fig:four_module_robot} and \ref{fig:applications}).

The main contributions of this work are as follows:
\begin{itemize}
\item A novel actuation method that utilizes compound folding and zipping of flexible 3D printed strips to enable continuous and reversible transformations in scale and stiffness

\item The design of a compact, minimal actuation module that can be easily integrated into a range of transformable and compliant robotic systems

\item Characterization of the actuator’s mechanical behavior across deployment states, including bending stiffness and buckling-induced failure modes.

\item Integration of multiple actuator modules into an adaptive quadruped robot, demonstrating basic walking, reaching and scale-changing maneuvers.
\end{itemize}

The remainder of this paper is organized as follows: Section II details the mechanical design and fabrication of the actuator; Section III presents mechanical characterizations and robotic demonstrations; and Section IV discusses limitations, scalability, and future research directions.

\section{Method: Design and Control}

\subsection{Overview}

\begin{figure}[t]
\centering
\includegraphics[width=1\linewidth]{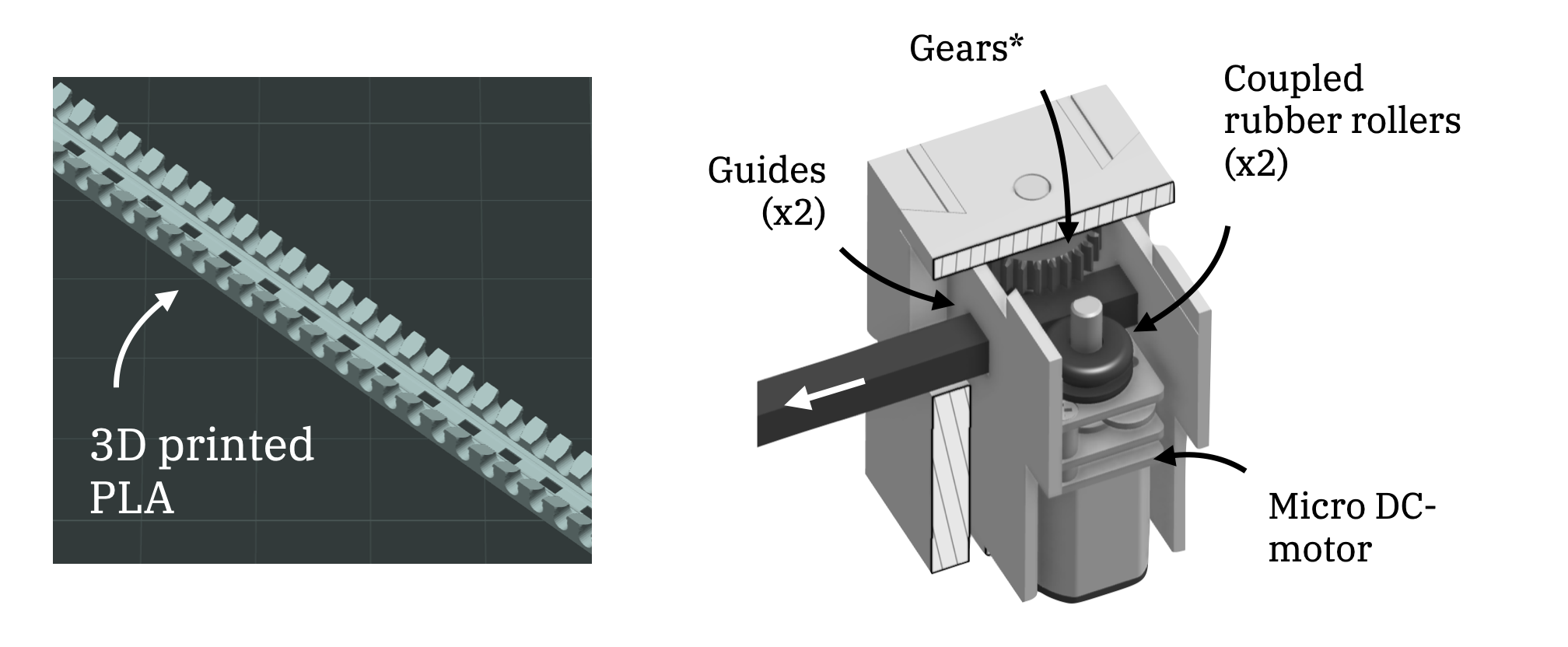}
\caption{\textbf{ZipFold module design and fabrication.} The ZipFold actuator makes use of the Fold-and-Zip principle illustrated in Figure \ref{fig:actuation_principle} and is quick to assemble from minimal components (a single geared DC motor, 3D printed gears and enclosure, a steel dowel pin and rubber rollers). The zipper strips are designed with oblique inwards-facing zipper features and a central hinge. They can be printed flat on a desktop 3D printer (e.g. Bambu X1, with a 0.4 mm nozzle) from PLA plastic.}
\label{fig:basic_concept}
\end{figure}

Our ZipFold actuator utilizes the geometry of folding to achieve stiffness by increasing the second moment of area of a flat normally flexible strip of material (see Figure \ref{fig:actuation_principle} and \ref{fig:basic_concept}). While this general principle has been used extensively for deployable boom systems \cite{miura_forms_2020}, these systems suffer under transverse (off-axis) loading conditions: with buckling and torsional compliance as dominant failure modes. With folding in one direction, there is a natural asymmetry (an up and a down), and this spills over into the bending stiffness. Booms used by NASA and others generally feature two opposing geometries for this reason \cite{crawford_strength_1971}\cite{thomson_deployable_1994} and researchers have also explored zipper-like interlocking of origami-inspired flat-folded booms \cite{kim_highly_2023} . However, these efforts generally come at the expense of mechanical complexity, additional motors, advanced fabrication processes involving composite materials and with buckling and torsional compliance still significant failure modes. We present our solution as a compact, simple and ultimately scalable alternative to the existing research that will facilitate integration into adaptive robotic systems more broadly.

\subsection{The Fold-and-Zip actuation principle}

The ZipFold actuator modules presented in this paper make use of a general deployment mechanism we will refer to as 'Fold-and-Zip'. The Fold-and-Zip principle is illustrated in Figure \ref{fig:actuation_principle}: two opposing flexible zipper strips are drawn together through a square guide 'window', and in one continuous step are folded and zipped into quasi-rigid square-section deployable beam.


\begin{figure}[t]
\centering
\includegraphics[width=.95\linewidth]{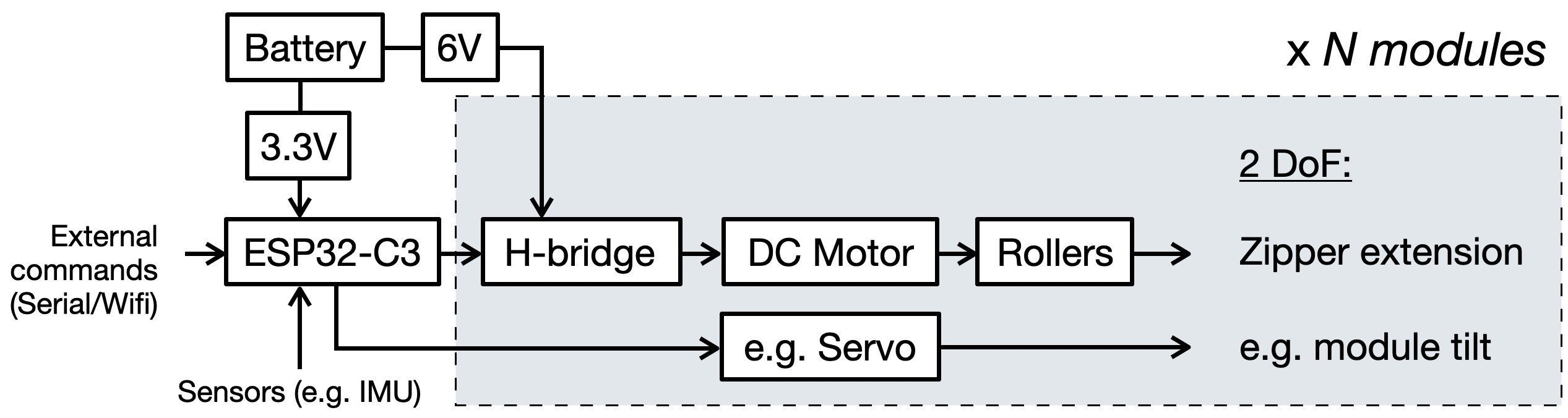}
\caption{\textbf{Control architecture for an n-module robot}}
\label{fig:generic_control}
\end{figure}

\begin{figure}[!t]
\centering
\includegraphics[width=.95\linewidth]{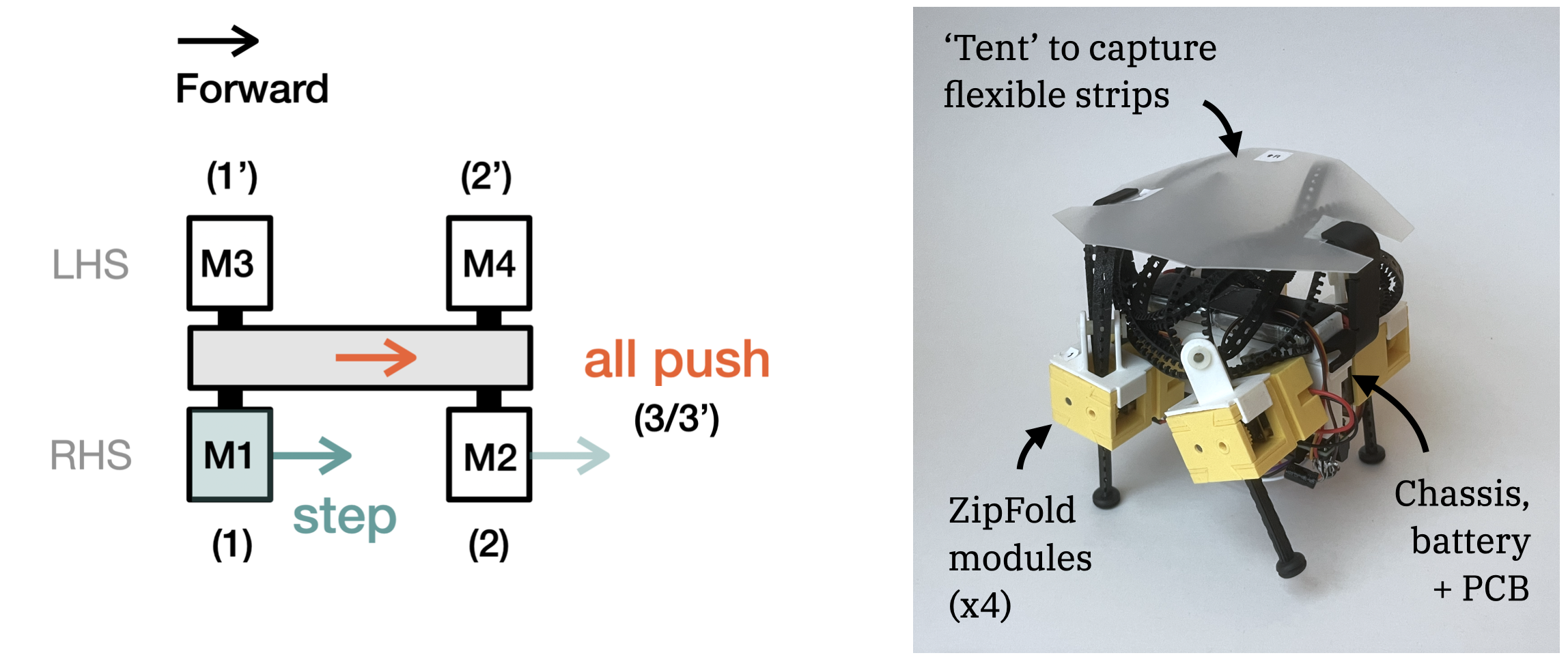}
\caption{\textbf{Overview of the four-module robot walker and crawl gait sequence.} Four ZipFold modules are mounted to a simple central chassis. The flexible zipper strips are 'collected' in an ad-hoc fashion at the top of the robot chassis. For locomotion, a simple 'crawl' gait is implemented. Each `step’ involves retracting an actuator, tilting the servo motor forward, and re-extending the zipper to a new position on the ground. `Push’ involves all modules pushing backwards.}
\label{fig:robot_walker_overview}
\end{figure}


\begin{figure*}[!t]
\centering
\includegraphics[width=0.95\linewidth]{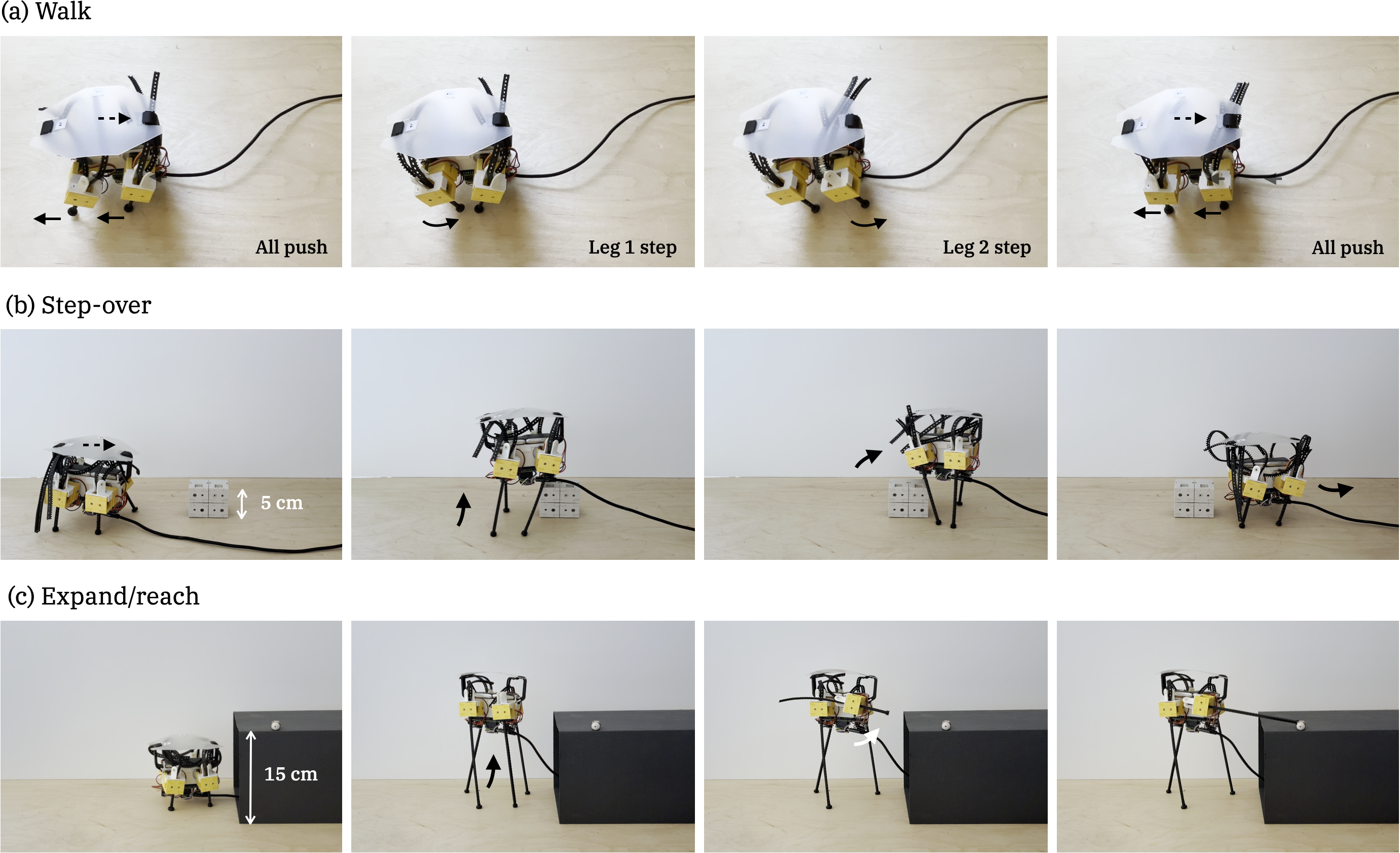}
\caption{\textbf{Experiments with the four-module walking robot} (a) A basic walk (compare to crawl gait sequence depicted in Figure \ref{fig:robot_walker_overview}), (b) stepping over a small obstacle (four stacked module shells, 4 x 6 x 5 cm), (c) expanding and reaching over an obstacle (15 cm height). A fourth experiment is shown in Figure \ref{fig:four_module_robot}, with the robot starting in an expanded state and crouching to crawl into a 15 cm pipe.}
\label{fig:experiments_combined}
\end{figure*}

\subsection{The ZipFold Module}

The implementation of the Fold-and-Zip principle in the form of a compact actuator module is illustrated in Figure \ref{fig:basic_concept}. A single ZipFold module measures 30 x 40 x 25 mm (w x d x h). Flexible strips are drawn into the module via two coupled counter-rotating rollers (i.e. driven by a single geared DC motor, see Figure \ref{fig:basic_concept}). The rollers perform dual functions: 'dragging' the flexible strips through the square guide cut-outs (see Figure \ref{fig:actuation_principle} and \ref{fig:basic_concept}), while simultaneously providing the clamping force required to press-zip the strips together. This simple, roller-based actuation scheme is fundamentally smooth, minimizing noise and reducing wear on the zipper strips.





Each zipper module consists of a single DC motor which is controlled by an H-bridge motor driver circuit, requiring two pulse-width modulated (PWM) signals from a micro-controller for speed and direction control.

\subsection{Integration of multiple modules}

Multiple ZipFold modules can be combined in series of parallel configurations to form soft or deployable robotic systems of varying scales and function (see Figure \ref{fig:applications}). At the current research stage, an adaptive four-legged (quadruped) robot has been built to demonstrate the capabilities for scale change and basic maneuvering (see Figure \ref{fig:four_module_robot}. For an added degree of freedom, a small position-controlled servo-motor has been integrated into each actuator module. This was achieved without any alterations to the module footprint, and could be seen as a standard addition to the ZipFold modules going forward.


\paragraph{Control}
A generic control architecture for a robotic system composed of an arbitrary number of ZipFold modules is shown in Figure \ref{fig:generic_control}. The current modules are operated in open-loop, with a single ESP32-C3 micro-controller controlling four ZipFold modules with integrated servo motors for an additional degree of freedom. An IMU can provide basic feedback for stabilization of roll and pitch of the robot via a simple PD control loop. Each module is controlled via an H-bridge driver circuit, allowing independent direction and speed control.

All components connect to a custom printed circuit board (PCB). In the future, the electronics could be further miniaturized such that each actuator module can be equipped with its own micro-controller and sensors, to enable more flexible distributed and reconfigurable systems.

\paragraph{Locomotion}
 To enable the four-module robot to walk, a `crawl' gait was chosen for it's stability and ease of implementation \cite{mcghee_stability_1968}. The two legs on each side of the robot alternate in making sequential steps, with  all legs pushing forwards after each side has completed its steps. The sequence is illustrated in Figure \ref{fig:robot_walker_overview}.


\section{Results}
To characterize the mechanical performance of the ZipFold system, a series of compression and bending tests were performed on universal testing machine (Instron). The different testing configurations and their implications for scaling of the ZipFold actuators are outlined below.

\begin{table}[tb]
\centering
\caption{\textbf{Overview: Metrics for the ZipFold actuator module}}
\label{tbl:zipper_metrics}
\begin{tabular}{lll}\toprule

\textbf{ZipFold Module}& & \\\midrule
 Footprint (w x d x h)& 3 x 4 x 2.5&cm\\
 Mass, actuator block& 28&g\\
 Mass, two zipper strips  (per 30 cm length)& 4&g\\
 Peak axial force at 28 cm extension& 12&N\\
\rowcolor[HTML]{FFFFFF} 
Current draw, unloaded extension @6V& 50& mA\\
 Extension speed @6V& 10&mm/s\\ \bottomrule 
\end{tabular}

\vspace{0cm}
\end{table}

\begin{table}[tb]
\centering
\caption{\textbf{Overview: Metrics for the four-module walking robot}}
\label{tbl:zipper_metrics}
\begin{tabular}{lll}\toprule
 \textbf{Four-module Robot} & & \\\midrule
 Footprint, compact (w x d x h)& 9 x 10 x 11&cm\\
 Mass& 270&g\\
 Standing height, compact & 11&cm\\
 Standing height, expanded & 32&cm\\
 Max reach (standing + one arm extended)& 55&cm\\
 Power draw, upwards expansion& 2.2&W\\ \bottomrule
\end{tabular}

\vspace{0cm}
\end{table}

\subsection{Scaling Laws}

\paragraph{Compression}

The most likely failure mode for a long slender beam (such as an extended zipper-actuator) under compressive loading is buckling. From Euler buckling theory \cite{budynas_shigleys_2015}, for a beam under axial loading with one end pinned, we know that:

\[F_e=C\frac{\pi^2EI}{l^2}\]

Where \(F_e\) is the critical Euler buckling load, \(E\) the Young's modulus of the material, \(I\) the second moment of area of the beam cross-section and \(C\) a constant dependent on the end-conditions of the other end of the beam. If we approximate the deployed actuator with \(I\) for a square cross-section of width \(a\), we can say that the buckling strength of the deployed zipper should scale with its deployed extension length, \(l\), by:

\[F_e \sim a^4/l^2\]

\paragraph{Bending}
Similarly to above, by considering normal stresses for a square section beam subject to an end-point bending load \cite{budynas_shigleys_2015}, we can show that the bending stiffness of the deployed zipper should scale via:

\begin{figure*}[!t]
\centering
\includegraphics[width=1\linewidth]{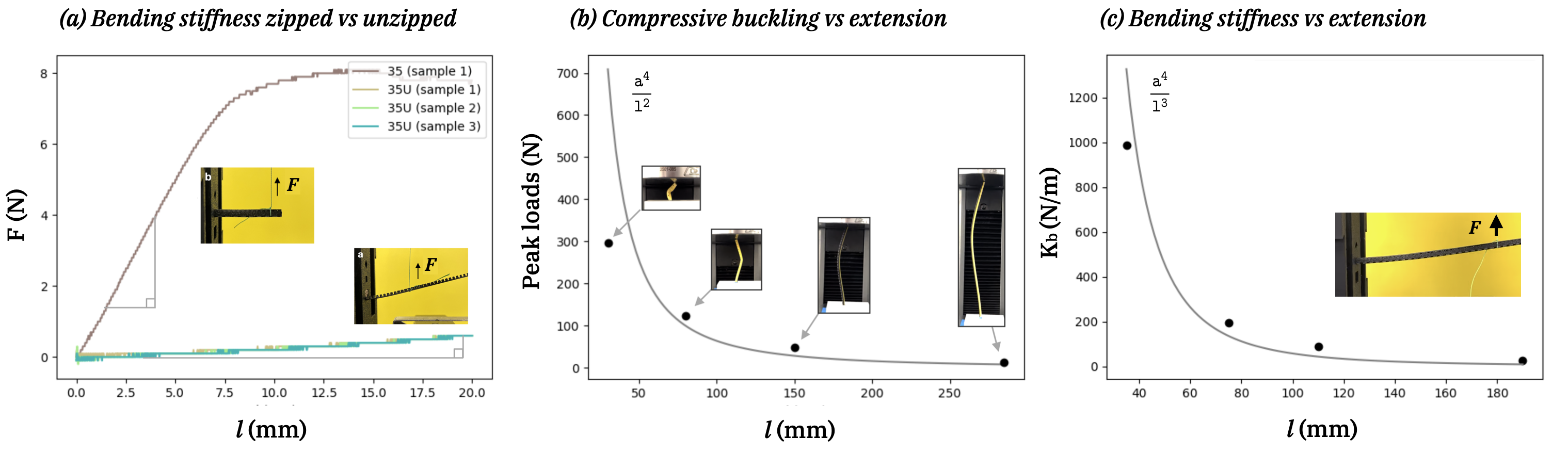}
\caption{\textbf{Mechanical characterization}. (a) Bending-stiffness tests for a single flexible strip and a deployed/zipped beam (a 36x increase in stiffness) (b) peak compressive loads at different deployment lengths, (c) bending stiffness for different deployment lengths.}
\label{fig:mechanical_characterization_plots}
\end{figure*}

\[K_b \sim a^4/l^3\]

\paragraph{Torsion}
Likewise, beam theory tells us that the torsional stiffness of the deployed zipper beam should scale according to:

\[K_\tau \sim a^4/l\]

\subsection{Mechanical characterization}

\paragraph{Compressive strength}
As expected, the compressive strength of the zipper rapidly decreases as it is extended, more or less with the inverse square of the extension (see Figure \ref{fig:mechanical_characterization_plots} (b)). As can be seen from the fitted scaling curve, the shortest deployment length (30 mm), while withstanding a significant peak load of ~ 30 kg, falls short of what we might expect from our idealized model. This is likely due to local stress concentrations (e.g. at the zipper notches) and 3D printing imperfections.







\paragraph{Bending stiffness}
Bending tests were performed at different extension lengths with a Nylon string and the beam cantilevered to the side (see Figure \ref{fig:mechanical_characterization_plots}) (a) and (c). Figure \ref{fig:mechanical_characterization_plots} (c) shows the computed bending stiffness of the deployed zipper at various extensions. \remove{One notable observation was the significant noise observed in the longer extensions (vs the shortest sample at 35 mm), perhaps again due to vibrations caused by localized zipper notches sliding against each other. Taking the gradient in the linear/elastic portions of the plots, we can estimate the bending stiffness.} As can be seen from the fitted curve, the reduction in stiffness with extension corresponds closely to the expected scaling law.

\paragraph{Stiffness-change}
To characterize the change in stiffness that occurs during deployment, a bending test was performed with a single flexible strip and a deployed two-zipper beam (see Figure \ref{fig:mechanical_characterization_plots} (a)). For the best force resolution, a short extension of 35 mm was chosen for the comparison. Comparing the gradient (in the elastic regions), we get a stiffness-change of ~ 36x between a single un-deployed zipper strip and the deployed zipped beam.

\paragraph{Isotropy}
To assess whether the bending stiffness of the deployed zipped beam was substantially affected by orientation, the beams were rotated +45 deg and -45 deg around their central axis. The bending stiffness in different orientations proved virtually the same, although pronounced noise appeared in the -45 degree rotation, perhaps due to unfavorable orientation of the zipped edges normal to the applied bending force causing sliding between zipper notches. Again, it is worth noting that unzipping itself never occurred as a failure mode.



\subsection{System}
ZipFold actuators can be composed into a range of adaptive systems. To demonstrate their capabilities for enabling shape and scale-change, a simple adaptive robot walker was composed of four modules.

The crawl gait proved an effective strategy: the back leg and then the front leg of one side make steps, followed by all four legs pushing forward. This is then repeated but with the other side making the steps, and so on (see Figure \ref{fig:robot_walker_overview} and \ref{fig:experiments_combined} (a)). The gait also worked in combination with other tactics like expansion. Figure \ref{fig:experiments_combined} (b) shows the four-module walker rising to pass over an obstacle 5 cm in height and 6 cm in length (and 4cm wide). The robot is also able to expand higher, rising to a maximum standing height of 32cm to reach over obstacles and extending actuator sideways across (see Figure \ref{fig:experiments_combined} (c)). Likewise, the robot is able to drop down from a taller, expanded position to become compact and crawl into a confined space (e.g. a pipe of height x width: 15 x 20 cm, see Figure \ref{fig:four_module_robot}). A limitation of the current robot is that the flexible zipper strips from different modules occasionally interlock with each other or get stuck on other parts of the assembly as the actuators are retracted and re-deployed. While the `ad-hoc' storage of the zipper strips can be beneficial in terms of packing efficiency, a more systematic/guided collection of the zipper strips would be beneficial to ensure robustness.

\section{Summary and future work}

We have presented the design for a compact deployable actuator based on the reversible folding and zipping of flexible 3D printed plastic strips into a square-section semi-rigid beam. The actuator is easy to fabricate using only a desktop 3D printer, minimal extra components and a small DC motor. The compact module footprint enables affordable and scale-able adaptive robotic systems that can transform their shape, scale and stiffness in response to changing environments or tasks.

The ZipFold actuator module demonstrate good dimensionality (straightness) even after many repeated deployments (zipping/unzipping) as well as predictable scaling in stiffness under bending and compression. Unzipping of the deployed beams was never observed as a failure mode during mechanical testing, suggesting that further improvements to the actuator mechanical strength should focus primarily on material selection (e.g. to counteract fatigue after repeated zipping/unzipping and excessive strains during coiling).

At higher extensions the actuators become prone to vibrations and reduced actuation precision which becomes problematic for robot stability. Future work should look into improved actuation methods, materials and zipper geometries to address this as well as adding closed-loop control schemes \cite{schneider_reachbot_2022}. In addition to inertial feedback, this might involve integrating pressure sensors at the tips of the deployed actuators and it would be interesting to look at sensing and control schemes that operate in a distributed fashion with many modules operating in parallel across a robot assembly.

The ZipFold modules have been integrated into a four-legged robot walker that is able to contract and expand to reach over and beneath obstacles. The flexible strips can be stored in an ad-hoc fashion, enabling compactness and allowing for the undeployed soft state of the actuators to be utilized: although future work should also consider more methodical storage solutions, to prevent friction or jamming between strips during deployment. Future work should look at integrating the modules into a wider range of soft- and soft-rigid hybrid systems, from novel shape-changing robots to lean assistive wearables and shape-changing interactive technologies\cite{hedayati_designing_2022}.

\section*{ACKNOWLEDGMENT}
The researchers would like to thank Zhenwei Ni at NUS for help conducting mechanical characterization experiments and Skylar Tibbits at MIT for helpful discussions. This research was supported by the Singapore-MIT Alliance for Research and Technology (SMART) M3S research grant.


\bibliographystyle{IEEEtran}
\bibliography{reference}






\end{document}